%% file: main.tex
\documentclass[letterpaper, 10 pt, journal, twoside]{ieeetran}
\usepackage{amsmath,amsfonts}

\usepackage{algorithm}
\usepackage{array}
\usepackage[caption=false,font=normalsize,labelfont=sf,textfont=sf]{subfig}
\usepackage{textcomp}
\usepackage{stfloats}
\usepackage{url}
\usepackage{verbatim}
\usepackage{graphicx}
\usepackage{cite}
\usepackage{xcolor}
\usepackage{hyperref}
\usepackage{algpseudocode}
\usepackage{booktabs}
\usepackage{amssymb}

\begin{document}

\title{Bilevel Planning with Learned Symbolic Abstractions from Interaction Data }

\author{Fatih Dogangun$^{1}$, Burcu Kilic$^{1}$, Serdar Bahar$^{1}$, and Emre Ugur$^{2,1}$%
\thanks{This work was supported by Scientific and Technological Research
Council of Turkey (TUBITAK) ARDEB 1001 Program (124E227) and by the
INVERSE Project under Grant 101136067, funded by the European Union.} 
\thanks{$^{1}$Department of Computer Engineering, Bogazici University,
Istanbul, Türkiye. $^{2} $Department of Computer Engineering, Middle East Technical University, Ankara, Türkiye.
        {Corresponding Author: \tt\footnotesize fatihdogangun16@gmail.com}}
\thanks {The code: \url{https://github.com/fatihdogangun/bilevel-planning.git} }
}

\maketitle

\begin{abstract}
Intelligent agents must reason over both continuous dynamics and discrete representations to generate effective plans in complex environments. Previous studies have shown that symbolic abstractions can emerge from neural effect predictors trained on a robot's unsupervised exploration. However, these methods rely on deterministic symbolic domains, lack mechanisms to verify the generated symbolic plans, and operate only at the abstract level, often failing to capture the continuous dynamics of the environment. To overcome these limitations, we propose a bilevel neuro-symbolic framework in which learned probabilistic symbolic rules generate candidate plans rapidly at the high level, and learned continuous effect models verify these plans and perform forward search when necessary at the low level. Our experiments on multi-object manipulation tasks demonstrate that the proposed bilevel planning method outperforms symbolic baselines, achieves the planning success of continuous forward search at a fraction of its computational cost, with a verification mechanism that reliably validates symbolic plans and guides the transition between planning levels.
\end{abstract}


\input{introduction}

\input{related_work}

\input{method}

\input{experiments}
\input{limitations}
\input{conclusion}

\bibliographystyle{IEEEtran}
\bibliography{ref}

\end{document}

%% file: introduction.tex
\section{Introduction}

Building robotic agents that can sense, plan, and act in continuous environments requires reasoning about the possible future states and effects of actions. However, reasoning directly in high-dimensional state space quickly becomes computationally intractable, since even simple tasks may have a large (or infinite) number of solutions in the continuous sensorimotor space of the robots. This motivates the need to learn high-level concepts that abstract task-related structures in the environment, thus reducing the complexity and enabling the use of efficient AI planners~\cite{konidaris2019necessity}. Beyond efficiency, such abstractions provide a mechanism for generalization, as once an environment is represented in an abstract model, high-level reasoning about transitions can be applied to different problem settings within that environment. In this regard, a key challenge is automatically acquiring abstractions that are both tractable to learn and expressive enough for planning.

There have been methods that rely on manually constructed state representations for planning; however, symbols defined this way are limited in their scope and scalability~\cite{ugur2025}. To obtain generalizable state abstractions, relational representations that capture object-object interactions (e.g., ``x on y'') are crucial. Early relational symbol learning approaches, such as~\cite{kulick2013active}, proposed a robot-teaching scheme where the robot actively generates different situations to learn grounded relational symbols from human labeling. In a more recent approach~\cite{shah2024reals}, inter-object representations were invented by capturing frequent relative states from raw demonstrations. However, these works required either human labeling or human demonstrations, which limit their scalability and autonomy. \cite{taniguchi2018symbol} argues that symbols can emerge from the robot's exploration within the environment. In this context,~\cite{ahmetoglu2025symbolic} utilized symbols that emerged from a bottleneck layer of an encoder-decoder-based effect predictor neural network. These studies have demonstrated success in high-level relational symbol emergence through the robot's random exploration of the state space, and planning with the learned symbolic transitions in an end-to-end fashion.

Although symbolic state abstractions enable efficient high-level reasoning in robotics, grouping diverse continuous states into discrete symbols causes inherent information loss~\cite{silver2022neurosymbolic} as the degree of state compression trades off with the capacity to represent finer characteristics of the state~\cite{abel2019state}. Therefore, the same action may produce different outcomes across states that share the same abstract symbol, a variation that deterministic operators cannot capture. This lossy, non-deterministic structure can be mitigated by probabilistic transition models that reflect the empirical distribution of the observed action effects~\cite{silver2021learning,pasula2007learning}. However, even with accurate transition probabilities, abstract symbols often fail to account for intricate geometric or dynamic constraints, leading to failed executions. In this regard, a verification method to confirm generated symbolic plans, and the use of a more accurate model that specializes in environment dynamics, enhancing the expressivity of the planning domain, can help mitigate the shortcomings of abstracting complex environments. This motivation suggests the use of a bilevel planning framework, which would combine the efficiency of symbolic planning at the high level with the precision of continuous search at the low level~\cite{silver2021learning, chitnis2022learning}. Within such a framework, continuous search through multi-step effect predictions can facilitate the verification of symbolic plans while providing a basis for low-level planning when symbolic reasoning alone is insufficient.

In this work, inspired by \cite{ahmetoglu2025symbolic}, we employ an encoder-decoder-based neural network architecture to learn state abstractions by predicting effects and extract symbolic operators from interaction data to perform symbolic planning using off-the-shelf AI models \cite{helmert2006fast}. While this approach enables learning predicates for efficient planning, it relies on deterministic symbolic domains, lacks a mechanism to assess whether the generated symbolic plan will succeed in the continuous domain, and provides no fallback when the symbolic plan fails to capture the true dynamics. To address these limitations, our contributions are as follows:

\begin{itemize}
    \item We propose a bilevel planning framework that encompasses planning at both symbolic and continuous levels, where continuous search serves as a fallback for problems that remain unsolved at the abstract level.
    \item We introduce a dual-model approach in which a symbolic model learns symbolic representations and probabilistic operators for high-level planning, while a forward dynamics model provides accurate effect predictions for plan verification and continuous forward search when no verified symbolic plan is available. 
    \item We introduce a verification mechanism that utilizes the continuous effect predictions to predict outcomes of candidate symbolic plans before execution, bridging the symbolic and continuous levels of the proposed framework.
\end{itemize}

We evaluate the proposed framework on multi-object manipulation tasks and show that it outperforms the prior approach, which relies solely on deterministic domains, and performs planning comparably to continuous forward search, while solving most problems efficiently through symbolic reasoning and resorting to more expensive search in the continuous state space only when required. We also show that using a probabilistic planning domain \cite{younes2004ppddl1} captures richer environmental behavior than a deterministic domain, improving the planning performance. Moreover, we validate the dual-model scheme by analyzing planning performance across both symbolic and continuous domains, and assess the verification mechanism's classification performance to demonstrate its reliability.

%% file: related_work.tex
\section{Related Work}

A large body of work focuses on state abstraction for planning \cite{ugur2025}. Constructing a symbolic domain requires learning the preconditions and effects associated with actions, as these enable the reasoning behind state transitions \cite{james2020learning, james2022autonomous, konidaris2018skills}. To this end, \cite{asai2018classical} compressed the continuous state space in an autoencoder module via unsupervised learning, then encoded the symbolic transitions to enable high-level planning. More recently, \cite{ahmetoglu2022deepsym} used effect-predictor networks to discover symbolic state representations, arguing that symbols should account for their prediction performance, and the symbolic transitions were used for probabilistic high-level planning. In follow-up work, \cite{ahmetoglu2024discovering} added self-attention layers to enable multi-object representations and learned relational symbols, while \cite{ahmetoglu2025symbolic} used these abstractions to generate symbolic transitions for planning in multi-object manipulation tasks.

Once symbolic representations are obtained, off-the-shelf planners can be used to perform efficient search. Many studies used frameworks such as Planning Domain Definition Language (PDDL)~\cite{ghallab1998pddl}, and its probabilistic extension Probabilistic Planning Domain Definition Language (PPDDL)~\cite{younes2004ppddl1}, classical planners such as Fast-Downward~\cite{helmert2006fast} for symbolic planning in both deterministic and stochastic domains.
In \cite{silver2021learning}, a bottom-up method for probabilistic operator learning was proposed, and the learned operators were used in task and motion planning (TAMP) systems. In \cite{chitnis2022learning}, relational transition models were learned to perform hierarchical planning consisting of symbolic and continuous levels. \cite{silver2022neurosymbolic,  li2025bilevel} learned predicates and proposed a framework to use these symbolic operators for bilevel planning; however, they used oracle demonstrators with manually defined abstractions to collect transition data. \cite{konidaris2015symbol} proposed a high-level planning with probabilistic effects by using a PPDDL domain, and \cite{ames2018learning} extended the probabilistic planning with symbolic representation learning by using parameterized skills. Recently, \cite{aktas2024vqcnmp} used vector quantization to cluster demonstrations into discrete skill representations, used in a bilevel pipeline where an LLM performs high-level planning. In this study, instead of providing demonstrations or foundation model inputs, we use the robot's own unsupervised interaction experience within a bilevel planning framework that leverages the efficiency of probabilistic symbolic planning and the accuracy of grounded continuous-state planning.

Since symbolic abstractions often discard geometric and dynamic information, several studies have focused on verifying symbolic plans or predicting their feasibility before execution. \cite{noseworthy2021active} proposed an active learning approach to acquire an abstract plan feasibility predictor through exploration on a robot. \cite{wells2019learning} learned a feasibility classifier for TAMP in tabletop environments. In a more recent work \cite{Yang2023PlanFeasibilityRSS}, transformers were leveraged for sequence-based prediction of feasibility in task and motion planning. 
Furthermore, \cite{liang2022search} trained skill-effect models and used them for continuous search via weighted A*. In our work, we train a forward dynamics model for effect prediction and use it for both plan verification and continuous forward search, invoked when symbolic plans are unverified.

%% file: method.tex
\section{Method}

In this section, we first outline the problem definition and provide background on unsupervised symbol discovery and symbolic rule learning, which form the basis for our method. We then present our framework, including the dual-model scheme, probabilistic symbolic planning, plan verification mechanism, and continuous forward search.

\subsection{Problem Definition}
\label{subsec:prob_def}

This study addresses the problem of efficient and accurate planning from discovered symbols and operators for robotic manipulation. While our bilevel approach entails different means to generate a successful plan, the main objective is to find a sequence of actions that brings a given initial configuration of objects into the desired goal configuration. The environment's continuous \textbf{state space} is denoted by $\mathcal{X}$. A specific state $X \in \mathcal{X}$ is represented as a set of feature vectors $\{x_1, ..., x_n\}$, where $x_i \in \mathbb{R}^{5}$ contains each object's 3D position and type (one-hot) in the scene. The agent can interact with the environment through a discrete set of high-level, parameterized skills, which constitute the \textbf{action space} $\mathcal{A}$. An action $a \in \mathcal{A}$ corresponds to a \texttt{pick-place} skill, defined by a tuple ($o_i,\delta_i,o_j,\delta_j$) where $o_i$ and $o_j$ indicate the picked and target objects, respectively, and $\delta_i$ and $\delta_j$ describe the relative positional offset for grasping and releasing, respectively. A \textbf{plan} is defined as a sequence of actions, $\pi\triangleq (a_1, a_2, \dots, a_k)$, where each $a_t \in \mathcal{A}$. Given a problem $P=(\mathcal{X}, \mathcal{A}, X_{init}, G)$, where $X_{init}$ is the initial state and $G$ is the goal condition, the task is to find a plan $\pi$ that results in a final state $X_{final}$ that satisfies the goal condition $G$. In this study, the goal condition is defined using a target state $X_{goal}$, and a plan is considered \textbf{successful} if the final state is within a minor tolerance, $\epsilon$, of this target: $\|X_{final} - X_{goal}\|_2 < \epsilon$.

\subsection{Background}
\label{subsec:preliminary}

Our framework builds upon the prior approach \cite{ahmetoglu2025symbolic}, which utilizes a deep learning model to discover both unary object and relational symbols from a dataset of interaction samples $(X, a, X')$, as illustrated on the left of Fig.~\ref{fig:dual_model_training}. At a high level, the learning procedure includes three stages: interaction samples are collected through random exploration, the network is trained on this collected data, during which object and relational symbols emerge in its bottleneck, and these symbols are used to form the symbolic operators used for planning. The primary objective during training is to predict the effect $e_i$ on each object $x_i$ by minimizing the mean squared error (MSE):

\begin{equation}
    L = \sum_{i=1}^{n} \left\| \hat{e}_i - e_i \right\|^2 
    \label{eq:loss}
\end{equation}
where $n$ is the number of objects, and $\hat{e}_i$ and $e_i$ are the
predicted and ground-truth effects of object $x_i$, respectively. The effect represents the change in position of objects subtracted by the lateral movement of the arm.

The model network contains two encoders: The object encoder, $\sigma_p$, is used to map the feature vector of each object $x_i$ to a unary symbol $\sigma_p(x_i) \in \{0,1\}^{d_z}$, and the relational encoder, $\sigma_{r}$, is used to compute binary relational symbols between each pair of objects. The relational encoder is an adaptation of the standard self-attention mechanism~\cite{vaswani2017attention}, utilizing the Gumbel-Sigmoid function, which allows the learning of discrete symbols in a differentiable manner. The relational symbol is computed as      $\sigma_{r}(x_{i},x_{j}) = \text{GumbelSigmoid}(q_{i} \cdot k_{j})$, and this technique allows the model to represent the relationships between a varying number of objects. The model then aggregates information using discovered symbols to predict the effect $e$ of an action $a$. The unary symbol $\sigma_p$ for each object's feature vector $x_i$ is combined with $a$ and is given as input to an MLP to produce an intermediate representation $z_i$. For each of the $K$ relation types (number of attention heads), the aggregation is computed as: $h_{i}^{k} = \sum_{j=1}^{n}\sigma_{r_{k}}(x_{i},x_{j})z_{j} \quad \forall k \in \{1, ..., K\}$. Then the resulting vectors are concatenated to form a complete representation for each object $h_i$, which is input to a decoder network that predicts the resulting effect $\hat{e}_i$.

The entire network is trained jointly and end-to-end by minimizing the MSE between the predicted and ground-truth effects (Eq.~\ref{eq:loss}). The discrete object and relational symbols are not supervised directly but emerge in the bottleneck layers as a result of optimizing effect prediction. Once trained, the model is used to extract symbolic rules, represented as a set of operators $\Phi = \{\phi_1, \dots, \phi_m\}$. To this end, the continuous transition dataset is first converted into a symbolic dataset. To achieve this, for a given state, the object and relational encoders first produce continuous activations in $[0,1]$, which are then discretized by rounding to the nearest integer (0 or 1) to obtain the binary unary and relational symbols. Applying this discretization procedure to every interaction $(X, a, X')$ converts the continuous transition dataset into a symbolic dataset consisting of pre-action states $\Sigma = (\Sigma_p, \Sigma_r)$ and post-action states $\Sigma' = (\Sigma_p', \Sigma_r')$. Transitions are then grouped by their symbolic pre-action states and action arguments, and the most frequent post-action state within each group is assigned as the resulting state of the operator. Following the PDDL action schema $\phi=(\text{pre}(\phi), \text{eff}(\phi))$, where $\text{pre}(\phi)$ denotes the precondition and $\text{eff}(\phi)$ the symbolic effect, the operators are defined as $\phi_a = (\Sigma, \Delta\Sigma)$ for each group with action arguments $a$. Here, $\Sigma$ is the group’s pre-action state, corresponding to the preconditions, and the symbolic effect,$\Delta\Sigma = (\mathcal{E}^+,\mathcal{E}^-)$, encodes the set differences between the pre-action state of the group and its post-action state. Operators are constructed for all such groups, yielding the complete set $\Phi$.

\subsection{Bilevel Planning with Continuous Verification}

\subsubsection{Overview}
To solve the planning problem defined in Section~\ref{subsec:prob_def}, we introduce a bilevel planning algorithm, which is described in Algorithm~\ref{alg:bilevel}.
The proposed algorithm is designed to improve the success and robustness of planning for manipulation tasks and operates at two levels. Level 1 comprises a probabilistic planning framework that captures a richer set of action outcomes compared to deterministic operators, bridging the gap between symbolic and continuous domains. Level 1 also includes a verification process to assess the plausibility of each candidate symbolic plan, thereby mitigating the need for trial-and-error and reducing the number of failed plan executions. If high-level reasoning fails to produce a verified plan, Level 2 utilizes a continuous forward search. The individual components of the proposed algorithm are described in detail in the following subsections. Overall, our bilevel approach improves planning performance compared to purely symbolic planners by incorporating continuous planning when necessary, while maintaining performance comparable to full continuous planning but at a reduced computational expense.

\begin{algorithm}[tb]
\caption{Bilevel Planning Algorithm}
\label{alg:bilevel}
\begin{algorithmic}[1]
\State \textbf{Input:} Initial state $X_{init}$, Goal state $X_{goal}$, Symbolic Model $M_{sym}$, Dynamics Model $M_{dyn}$, Symbolic Domain $\Phi$
\State \textbf{Output:} A plan $\pi$

\State {\textbf{Level 1:}}
\State {\textit{--- Probabilistic Symbolic Planning}}
\State $\Sigma_{init}, \Sigma_{goal} \leftarrow \text{SymbolicProblem}(X_{init}, X_{goal}, M_{sym})$
\State $\mathcal{D} \leftarrow \text{GenerateSampledDomains}(\Phi, N)$
\State $\Pi_{sym} \leftarrow \text{ProbabilisticSymbolicPlanner}(\Sigma_{init}, \Sigma_{goal}, \mathcal{D})$

\State {\textit{--- Verification}}
\For{$\pi_{sym} \in \Pi_{sym} $ in decreasing order of probability} 
    
    \State $\hat X_{final} \leftarrow \text{ContinuousPlanVerifier}(\pi_{sym}, X_{init}, M_{dyn})$
    \If{$\| \hat X_{final} - X_{goal} \|_2 < \tau_{verify}$}
        \State \Return $\pi = \pi_{sym}$ 
    \EndIf
    
\EndFor
\State {\textbf{Level 2:}\textit{ Continuous Search Fallback } }
\State $\pi_{cts} \leftarrow \text{ContinuousForwardSearch}(X_{init}, X_{goal}, M_{dyn})$
\State \Return $\pi=\pi_{cts}$ 

\end{algorithmic}
\end{algorithm}

\subsubsection{Dual-Model Scheme}

\begin{figure}[t]
    \centering
    \includegraphics[width=0.86\columnwidth]{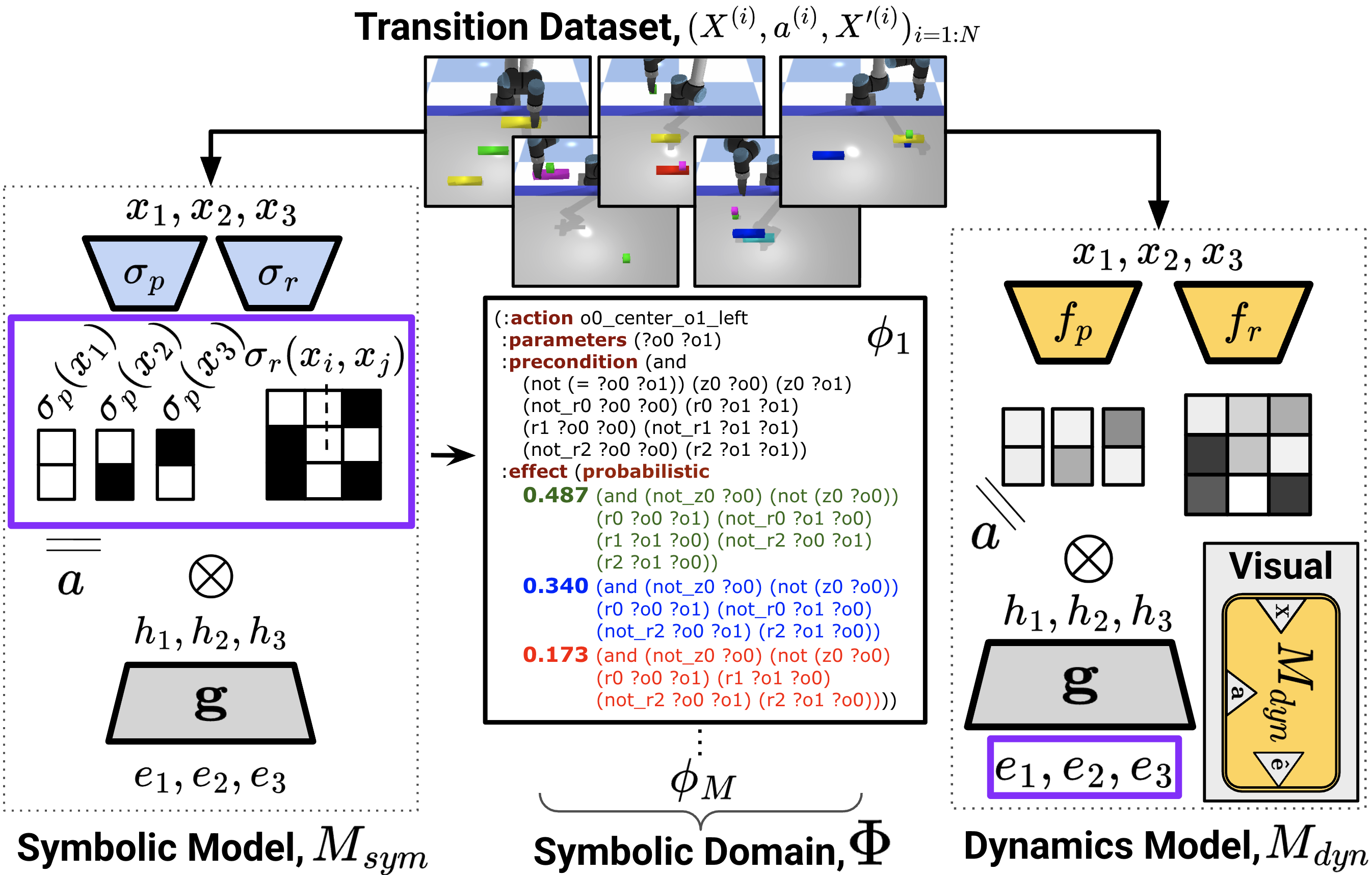}
    \caption{\textbf{Learning pipeline of the framework.} The system learns a \textbf{Symbolic Model} ($M_{\text{sym}}$) and a \textbf{Dynamics Model} ($M_{\text{dyn}}$) from a transition dataset of interaction samples. The \textbf{Symbolic Model} is used to generate symbols and find rules to form a symbolic domain $\Phi$ of probabilistic operators. The \textbf{Dynamics Model} is used for accurate effect prediction to verify symbolic plans and to search in continuous state space when no verified symbolic plan is available.
    }
    \label{fig:dual_model_training}
\end{figure}

A single model struggles to balance the discretization required for symbolic reasoning with the precision for planning in continuous state space.
Therefore, our framework utilizes the architecture described in Section~\ref{subsec:preliminary} and extends it with a continuous variant, forming the dual-model scheme as illustrated in Fig.~\ref{fig:dual_model_training}.

The \textbf{Symbolic Model} uses \textbf{Gumbel-Sigmoid} as an activation function, and its purpose is to discover symbolic state representations required for generating symbolic planning operators. The \textbf{Dynamics Model} has the identical architecture but utilizes \textbf{Sigmoid} activation in the bottleneck layers. This configuration enhances the capability for continuous effect prediction, positioning it as an optimal option for tasks necessitating a forward dynamics model. Although a Dynamics Model does not require this bottleneck architecture, we deliberately reuse the Symbolic Model’s architecture and modify only the activation function in the bottleneck layers to ensure two models remain directly comparable and a single shared architecture can serve both planning levels. We validate the activation choice empirically.

Both Symbolic and Dynamics models are trained independently as two separate networks with no shared parameters, and both are optimized under the same objective, the effect-prediction loss of Eq.~\ref{eq:loss}. There is no symbol-level supervision or auxiliary loss for either model, and each is optimized solely to minimize the effect prediction error. 

\subsubsection{Probabilistic Symbolic Planning}
\label{subsec:prob_plan}
\begin{figure}[t]
    \centering
    \includegraphics[width=0.76\columnwidth]{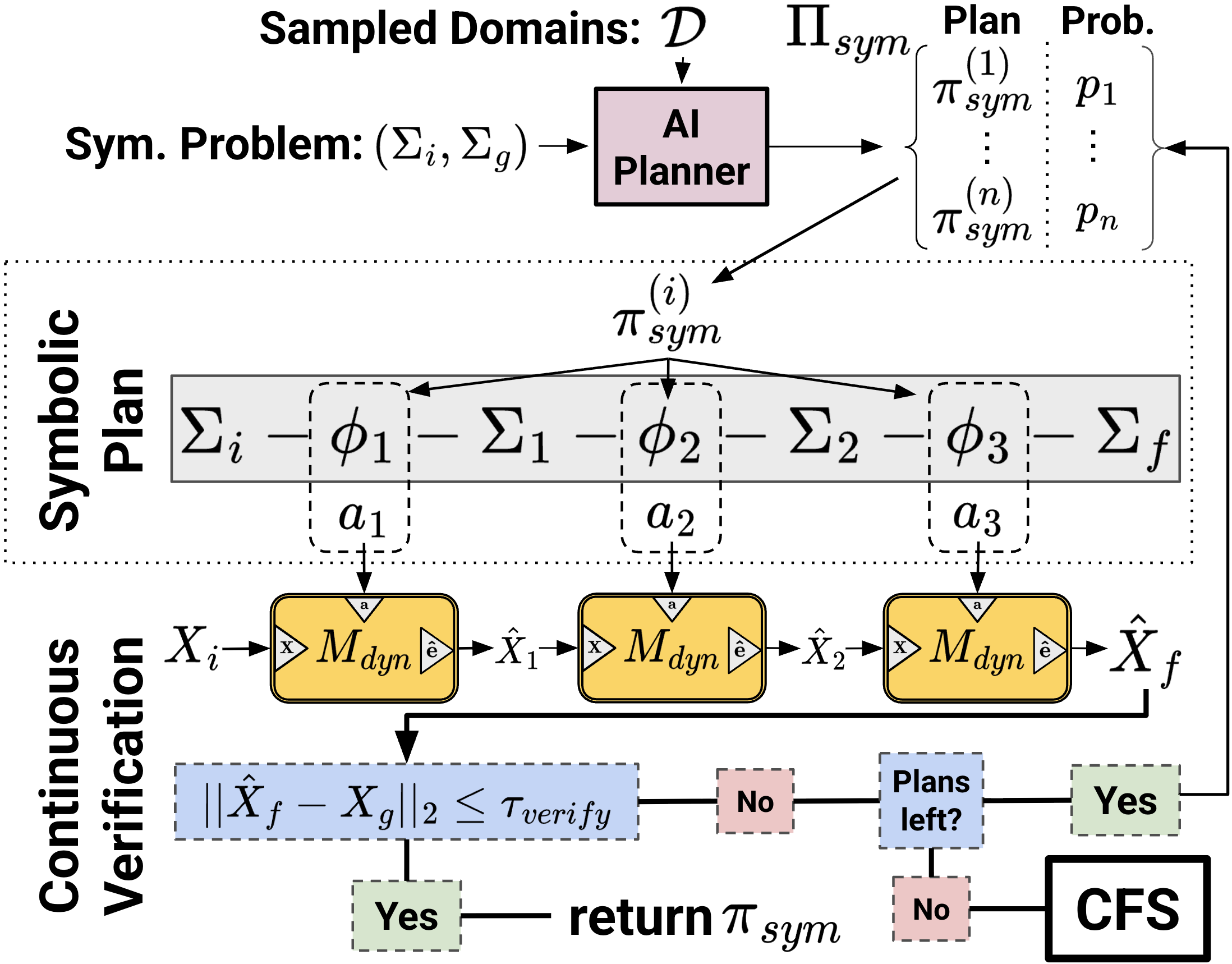}
    \caption{\textbf{Probabilistic symbolic planning and verification.} PDDL domains, $\mathcal{D}$, are sampled from the symbolic domain $\Phi$, and given to an AI planner with a symbolic problem. Then, candidate plans are verified in order of occurrence, and each action is checked in the continuous domain using the \textbf{Dynamics Model}. If a plan passes verification within a tolerance $\tau_{\text{verify}}$, the algorithm returns that symbolic plan; otherwise, it invokes a continuous forward search.
    }
    \label{fig:prob_verify}
\end{figure}

To improve planning success, we explicitly address the stochasticity of the learned operators, which is captured by the PPDDL domain. Our approach for probabilistic operator learning and planning is illustrated in Fig.~\ref{fig:dual_model_training} and Fig.~\ref{fig:prob_verify}. Operators that only consider the most probable outcome of an action can be brittle, since less likely but significant effects may be ignored yet prove crucial for a successful plan. Therefore, as illustrated in the center of Fig.~\ref{fig:dual_model_training}, we define a probabilistic operator, $\phi$, where the effect is a probability distribution over a finite set of possible symbolic effects, $E = \{\Delta\Sigma_1, \Delta\Sigma_2, \dots, \Delta\Sigma_m\}$, based on their empirical frequency: $\text{eff}(\phi) = \{ (\Delta\Sigma_j, p_j) \}_{j=1}^{m}$, where $p_j = P(\Delta\Sigma_j | \text{pre}(\phi), a)$ is the empirical frequency of the effect $\Delta\Sigma_j$, and $\sum_{j=1}^{m} p_j = 1$.

To perform probabilistic planning with a classical planner, we employ a sampling-based projection strategy. We construct an ensemble of $N$ PDDL domains (Alg.~\ref{alg:bilevel}, line 6), $\mathcal{D} = \{D_1, D_2, \dots, D_N\}$, where each domain $D_k$ is created by sampling a single effect for each probabilistic operator according to the corresponding learned probability distribution. Upon generating $N$ domains, we invoke the classical AI planner~\cite{helmert2006fast} (Alg.~\ref{alg:bilevel}, line 7) to search for a plan in each generated domain, allowing the system to explore a wider range of potential paths, and thereby increasing the robustness and success rate of symbolic planning. Then, we record each unique plan along with its occurrence among the generated plans, and use the normalized occurrence counts to define the probability distribution over the candidate plans. Consequently, as shown in the upper row of Fig.~\ref{fig:prob_verify}, we prioritize the most frequently discovered symbolic plans during the subsequent verification step, since they have the highest likelihood of success.

\subsubsection{Continuous Plan Verification}
\label{subsec:plan_verify}
To increase the success rate of execution of symbolic plans, we propose a verification mechanism (Alg.~\ref{alg:bilevel}, line 10) that leverages the continuous effect prediction capability of the \textbf{Dynamics Model}. Here, we define a \textbf{verifier} as a mapping $\nu: \Pi \times \mathcal{P} \rightarrow \{ \texttt{Verified},\texttt{Unverified} \} $, where $\Pi$ is the set of plans, and $\mathcal{P}$ is the set of problems. In our setting, a perfect verifier would evaluate a plan as \texttt{Verified} when $||X_{final} - X_{goal}||_2 < \epsilon$.

Given a candidate symbolic plan, $\pi_{sym} = (a_1, \dots, a_k)$, we use the trained \textbf{Dynamics Model} as a forward dynamics model. As illustrated in Fig.~\ref{fig:prob_verify}, starting from the initial continuous state $X_{init}$, we simulate the plan by iteratively predicting the continuous effect of each action, and computing the new state by adding the predicted effect to the current state until we reach the predicted final continuous state $\hat X_{final}$. We approximate the verifier by considering a symbolic plan as verified if the predicted final state is within the goal tolerance: $|| \hat X_{final} - X_{goal}||_2 < \tau_{verify}$. As shown in Fig.~\ref{fig:prob_verify} (also Alg.~\ref{alg:bilevel}, lines 9-14), the verification proceeds until a plan is found that satisfies the verification threshold or all plans are exhausted. In general, the described verification process plays a critical role in increasing the reliability and robustness of the system by eliminating inconsistent symbolic plans before execution.

\subsubsection{Continuous Forward Search}
\label{subsec:cts_search}

For problems where symbolic planning is insufficient, we introduce a continuous forward search (Alg.~\ref{alg:bilevel}, line 16) as a fallback mechanism. We implement the search as a weighted $A^*$ algorithm that operates directly in the continuous state space $\mathcal{X}$. A node $s$ in the search tree is defined by a tuple $(X_s,g(s),\pi_s)$, and the evaluation function is expressed as: $f(s) = g(s) + w*h(X_s)$, where $g(s)$ is the cost of the plan, $h(X_s)$ is the heuristic function, and $w$ is the heuristic weight (set as 1.5 in our experiments). The heuristic is defined as the sum of Euclidean distances between the positions of $n$ objects in the current and goal states: $h(X_s) = \sum_{i=1}^{n} \left\| \operatorname{pos}_i\!\left(X_s\right) - \operatorname{pos}_i\!\left(X_{\mathrm{goal}}\right) \right\|_{2}$. The predicted successor state,  $\hat X'$, for an expansion is generated using the \textbf{Dynamics Model} as the transition function, $\hat X' = X + \hat{e}$, where $\hat{e}$ is the predicted effect.

We also introduce heuristic action pruning to enhance the tractability of the continuous search mechanism. We generate a finite set of goal-oriented actions in each state, $\mathcal{A}(X_s)$, instead of exploring the entire action space, which contains $9n^2$ high-level parameterized actions for $n$ objects. First, we compute the 2D position displacement of objects between the current state $X_s$ and the goal state $X_{goal}$. We then identify the misplaced objects, $O_{misplaced}$, and generate two types of candidate actions for each $o_i \in O_{misplaced}$. First, we find the target object $o_j$ that is currently closest to the position of $o_i$ in the goal state $X_{goal}$. Then, generate actions from $o_i$ to $o_j$ with the full range of grasp and release offsets, which yield $9n$ actions in the worst case. Second, we generate actions from $o_i$ to itself, using all combinations of offsets, yielding $9n$ actions in the worst case. This pruning strategy reduces the branching factor of the search tree as the number of objects increases, making $A^*$ search computationally feasible and effective.

%% file: experiments.tex
\section{Experiments}
\label{sec:experiments}

\subsection{Experimental Setup}
\subsubsection{Environment}
We conducted experiments in a simulated tabletop environment, as shown in Fig.~\ref{fig:dual_model_training}. The environment contains a varying number ($n \in \{{2,3,4}\}$) of large and small blocks and a UR10e manipulator. To generate planning problems, we first created an initial scene $X_{init}$, where the types and positions of objects were randomly chosen. Then, we executed a sequence of $k$, where $k \in \{{1,2,3,4,5}\}$, random high-level actions to generate a goal state $X_{goal}$.

\subsubsection{Training}

We trained both models for 4000 epochs on $160$k interaction samples, using Adam optimizer~\cite{kingma2014adam} (learning rate: $0.0001$, batch size: $128$). Both models follow the architecture described in Section~\ref{subsec:preliminary}, differing only in activation functions: a Gumbel-Sigmoid with temperature set to $1.0$ for the \textbf{Symbolic Model} and Sigmoid for the \textbf{Dynamics Model}.

\subsubsection{Planning Methods}

We evaluate the performance of the following methods to analyze the contributions of our proposed framework: \textbf{Deterministic Symbolic Planner (Baseline):} The planner uses the PDDL domain, representing the baseline approach from~\cite{ahmetoglu2025symbolic}; \textbf{Probabilistic Symbolic Planner:} The proposed probabilistic planner approach that searches through an ensemble of $N$ PDDL domains from the PPDDL domain; \textbf{Continuous Forward Search:} The weighted $A^*$ search, finding plans in continuous state space; \textbf{Bilevel Planner (Proposed):} Our bilevel algorithm that attempts to plan with the Probabilistic Symbolic Planner with verification, and then invokes Continuous Forward Search if the first level fails. We used the Fast Downward~\cite{helmert2006fast} for symbolic planning. For each combination of object and action counts, we used 100 unique problems, with planning success rate as our primary metric. A trial succeeds if the executed plan yields a final state in which the Euclidean distance of all objects from their goal positions is below the tolerance threshold $\epsilon = 5$ cm. All reported significance values are obtained from paired t-tests.

\subsection{Results \& Discussion}
\label{subsec:results}

\subsubsection{Validation of the Dual-Model Design}
\label{subsubsection:model_performance}

\begin{table}[tb]
\centering
\caption{Mean Effect Prediction Error (cm) Across Bottleneck Activations Over Ten Runs}
\label{tab:eff_pred}
\begin{tabular}{cccc}
\toprule
 \textbf{Gumbel-Sigmoid} & \textbf{Tanh} & \textbf{No activation} & \textbf{Sigmoid} \\
\midrule
 $3.34 \pm 0.18$ & $2.24 \pm 0.12$ & $2.19 \pm 0.08$ & $\mathbf{2.01 \pm 0.13}$ \\
\bottomrule
\end{tabular}
\end{table}

We use two specialized models: a \textbf{Symbolic Model} for discovering symbolic representations and rules, and a \textbf{Dynamics Model} for tasks that involve continuous prediction. To validate the proposed dual-model approach, we first assess the activation function in the \textbf{Dynamics Model} based on effect prediction performance on a test set of 20,000 transitions. As shown in Table~\ref{tab:eff_pred}, the Gumbel-Sigmoid activation yields the highest effect-prediction error, motivating a \emph{different} activation for the \textbf{Dynamics Model}. Therefore, we further ablate the bottleneck activation and compare Sigmoid, Tanh, and unbounded (no-activation) variants. The Sigmoid bottleneck yields the lowest effect-prediction error, demonstrating its suitability as an activation function for reasoning in continuous state space.

\begin{figure}[t]
    \centering
    \includegraphics[width=0.94\columnwidth]{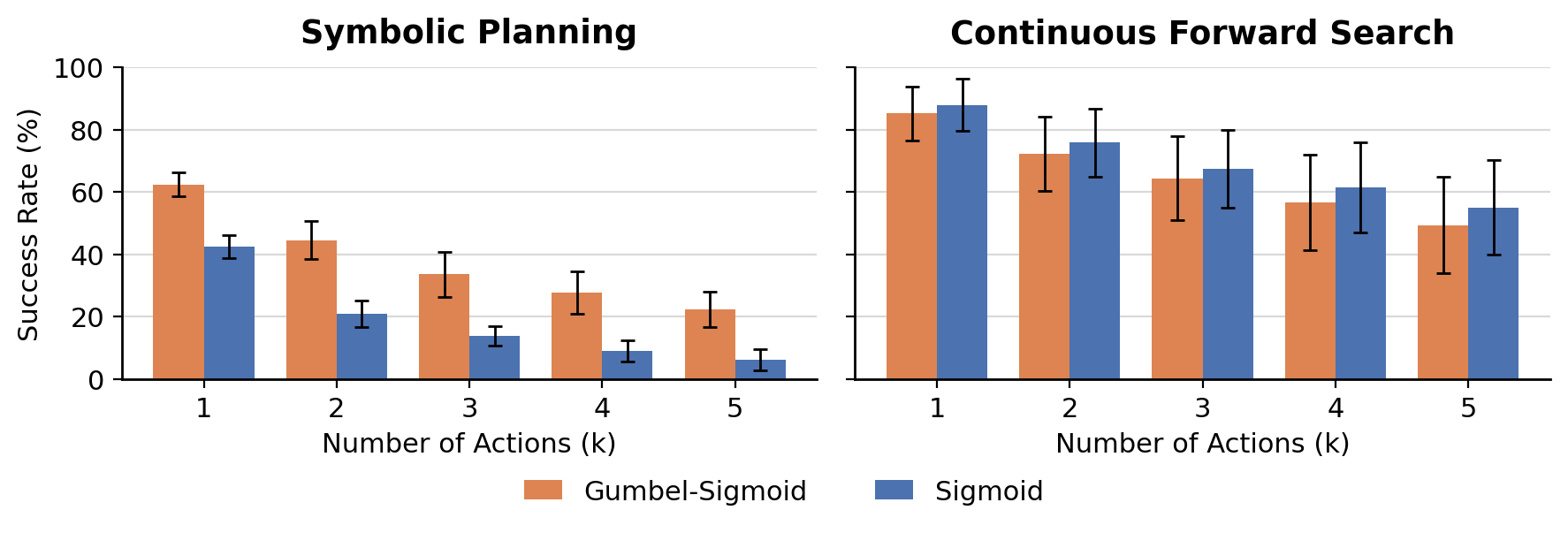}
    \caption{Planning success rate comparison of the Gumbel-Sigmoid and Sigmoid variants of the \textbf{Symbolic Model} (in symbolic planning) and the \textbf{Dynamics Model} (in continuous forward search). Results are averaged over ten runs and the number of objects ($n \in {\{2,3,4\}}$).}
    \label{fig:dual_model}
\end{figure}

Additionally, we compared how different bottleneck activations affect downstream planning performance in both the symbolic domain and the continuous state space. For symbolic planning, we evaluated two variants of the \textbf{Symbolic Model} where Gumbel-Sigmoid and Sigmoid activations are used in the bottleneck layers. Both variants generate a symbolic domain by discretizing their bottleneck layers to obtain state symbols, and the resulting abstractions are used to construct a symbolic transition dataset and extract operators (Section~\ref{subsec:preliminary}), after which Fast Downward is invoked to obtain plans. For continuous forward search, similarly, we evaluated the Gumbel-Sigmoid and Sigmoid variants of the \textbf{Dynamics Model} where both variants predict continuous effects, and their outputs are directly used to construct successor states as follows: $\hat X' = X + \hat{e}$. Fig.~\ref{fig:dual_model} reports all evaluated variants and shows the performance trade-off between different choices of bottleneck activations. While planning with learned operators in the symbolic domain, the Gumbel-Sigmoid variant of the \textbf{Symbolic Model}, significantly outperforms the Sigmoid variant ($p< 0.0001$). The Gumbel-Sigmoid activation biases the bottleneck layer toward binary representations during training, which results in more consistent symbolic operators compared to the Sigmoid activation, leading to better performance in symbolic planning. On the other hand, the Sigmoid variant of the \textbf{Dynamics Model} performs significantly ($p< 0.0001$) better planning in the continuous state space than the Gumbel-Sigmoid variant, with its superior predictive performance. As shown in Table~\ref{tab:eff_pred}, the Sigmoid variant achieves a lower effect prediction error, which yields more precise successor state predictions during the A* expansion, directly creating an advantage in planning in continuous state space. Using the Gumbel-Sigmoid activation variant for both the \textbf{Dynamics Model} and the \textbf{Symbolic Model} as a single-model scheme is feasible, but it reduces continuous planning performance.  Consequently, we adopted the dual-model scheme that uses the best-performing variant for each task: Gumbel-Sigmoid for the \textbf{Symbolic Model} and Sigmoid for the \textbf{Dynamics Model}.

\subsubsection{Planning Performance Comparison of Deterministic and Probabilistic Planning}

To evaluate the effectiveness of the probabilistic planning, which uses the full distribution of effects, we compared it to baseline deterministic planning that uses only the most probable effect. We used $N = 100$ sampled domains, and Table~\ref{tab:ppddl} shows that the ability to explore plans with less frequent effects in addition to the most frequent effect leads to a significantly ($p< 0.0001$) higher success rate.

\begin{table}[t]
\centering
\caption{Planning Success Rate (\%) comparison of deterministic and probabilistic planning. The results are averaged over ten runs and the number of objects ($n \in {\{2,3,4\}}$).}

\label{tab:ppddl}
\begin{tabular}{@{}ccc@{}}
\toprule
\textbf{Actions}& \textbf{Deterministic Planner} & \textbf{Probabilistic Planner} \\ 
\midrule
\textbf{k = 1} & $62.40 \pm 3.86$ & $\mathbf{71.70 \pm 3.48}$ \\
\textbf{k = 2}& $44.57 \pm 6.16$ & $\mathbf{54.87 \pm 5.07}$ \\
\textbf{k = 3} & $33.60 \pm 7.20$ & $\mathbf{43.90 \pm 6.57}$ \\
\textbf{k = 4} & $27.63 \pm 6.75$ & $\mathbf{35.67 \pm 6.81}$ \\
\textbf{k = 5}& $22.27 \pm 5.71$ & $\mathbf{28.93 \pm 5.56}$ \\
\bottomrule
\end{tabular}
\end{table}

\begin{figure}
    \centering
    \includegraphics[width=0.88\columnwidth]{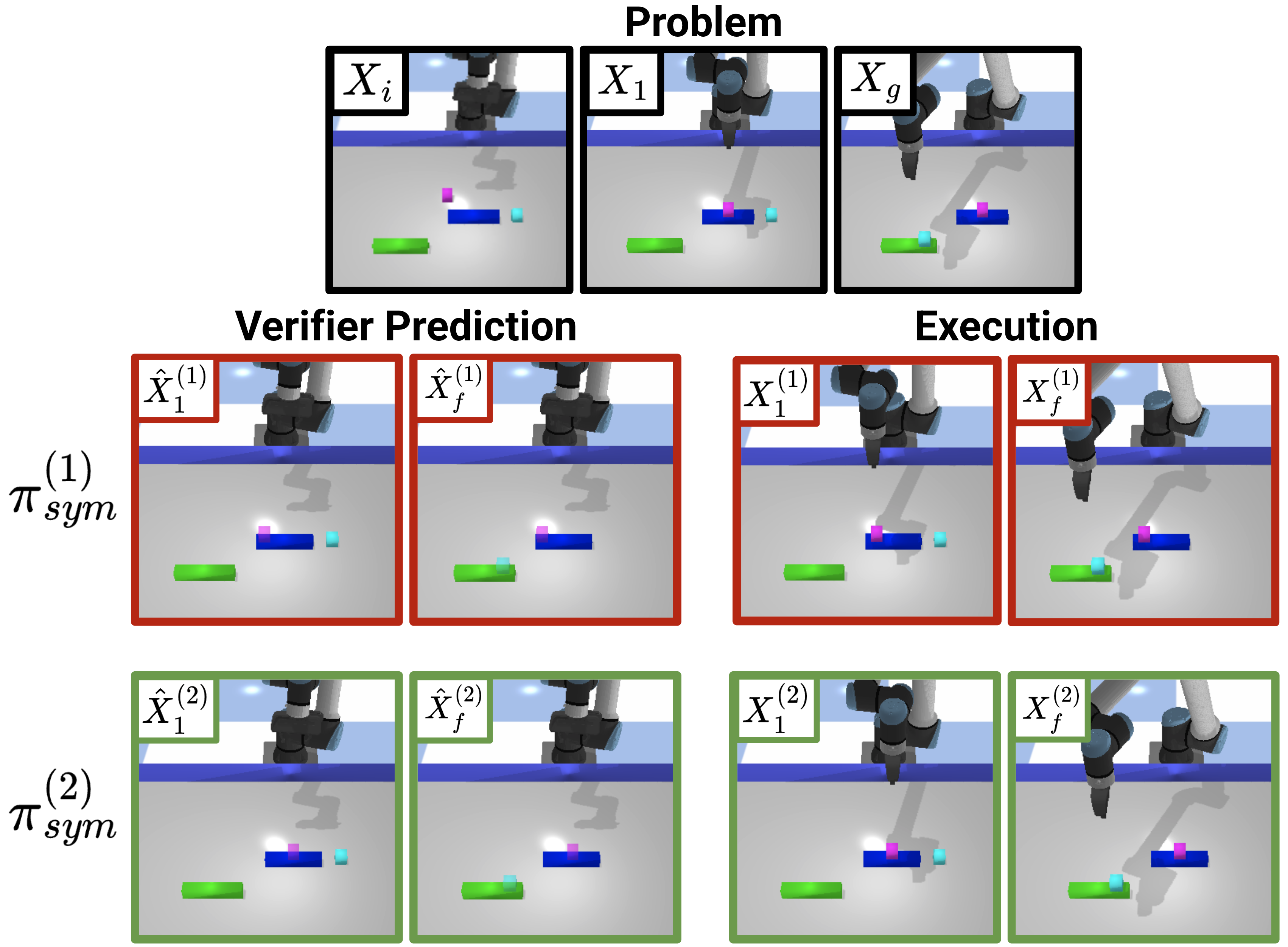}
    \caption{The top row shows the initial and goal states for the problem. The following rows show the sequence of plan verifier-predicted states (transparent objects illustrate the predicted positions) and their executions. The middle row presents the most probable plan, which also corresponds to the only symbolic plan if a probabilistic planner is not employed. The \textbf{Dynamics Model} accurately predicted the misalignment of the pink block's placement and classified the plan as \textbf{unverified}. The bottom row shows the next symbolic plan, which has been \textbf{verified} and deemed successful. 
}
    \label{fig:simulated_plans}
\end{figure}

The construction of deterministic operators is inherently limited, as it considers only the most frequent symbolic effect for a given set of preconditions while discarding all other observed outcomes, which leads to reduced expressivity and the potential loss of solutions in some cases. Object relations are not learned perfectly, and consequently, they do not transfer flawlessly into the symbolic domain, resulting in imperfect operators. Moreover, randomly collected training data includes an unequal distribution of transitions, which can bias learned operators toward specific scenarios. Probabilistic planning helps address these challenges by leveraging the full distribution of outcomes, allowing the system to solve problems where the most probable plan fails, as illustrated in Fig.~\ref{fig:simulated_plans}, ultimately leading to improved planning performance.

\subsubsection{Effectiveness of the Plan Verification Mechanism}
\begin{figure}[tb]
    \centering
    \includegraphics[width=0.95\columnwidth]{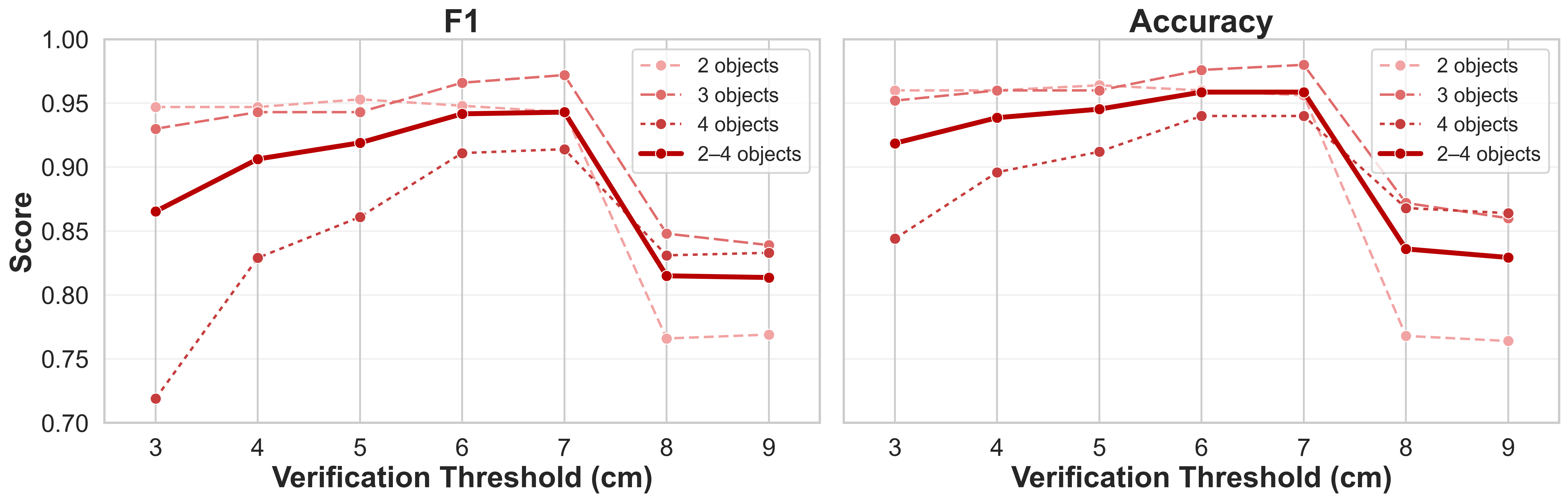}
    \caption{The F1 and Accuracy scores of the verifier for symbolic plans across varying thresholds ($\tau_{verify}$), over 50 problems, and the high-level actions ($k \in {1,2,3,4,5}$) for each number of objects.} 
    \label{fig:verify_threshold}
\end{figure}

To evaluate the verification mechanism, we analyzed the impact of the verification threshold $\tau_{verify}$ on the classification performance of symbolic plans. As shown in Fig.~\ref{fig:verify_threshold}, the choice of $\tau_{verify}$  strongly affects the performance. For larger $\tau_{verify}$, problem cases with a low number of objects were excessively classified as successful, since these simpler problems were already predicted well. On the other hand, low $\tau_{verify}$  values cause the verifier to perform poorly on tasks with more objects, since the problems are more challenging and require a greater margin of error. Therefore, we selected a $\tau_{verify}$ of $7 \,\mathrm{cm}$, resulting in the most consistent performance across varying numbers of objects. An illustrative example of the verification mechanism used for two symbolic plans is shown in Fig.~\ref{fig:simulated_plans}.

\begin{figure}[t]
    \centering
    \includegraphics[width=0.93\columnwidth]{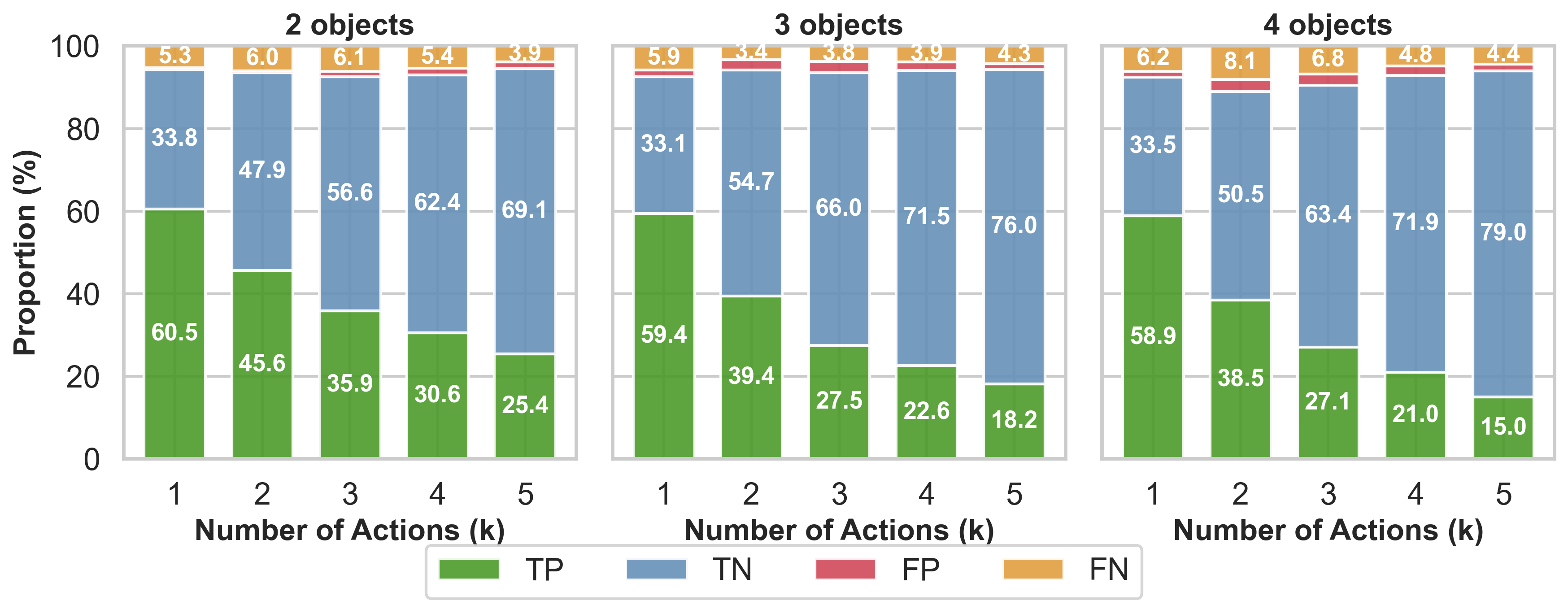}
    \caption{Continuous Plan Verifier performance across object and action counts, given that ($\tau_{verify} = 7\mathrm{cm}$). Plans are classified as: True Positive (TP): verified and successful, False Positive (FP): verified and failed, True Negative (TN): unverified and failed, False Negative (FN): unverified and successful.}
     
    \label{fig:verify_results}
\end{figure}

Fig.~\ref{fig:verify_results} shows the classification performance of the Continuous Plan Verifier with the chosen $\tau_{verify}$. Across all test scenarios, the symbolic plans that passed the verification step achieved a success rate of 94.2\%, while symbolic plans that were identified as unverified had a failure rate of 90.9\%, and the verifier achieved a mean accuracy of $0.93$ and a mean F1 score of $0.90$, confirming that the \textbf{Dynamics Model} serves as an effective mechanism to simulate plans and anticipate outcomes. Consequently, the Continuous Plan Verifier enhances the overall reliability by ensuring that the symbolic planner's output is grounded in the learned continuous dynamics.

\subsubsection{Planning Performance of the Bilevel Framework}
\label{subsec:overall}

\begin{figure}[tb]
  \centering
  \includegraphics[width=0.93\columnwidth]{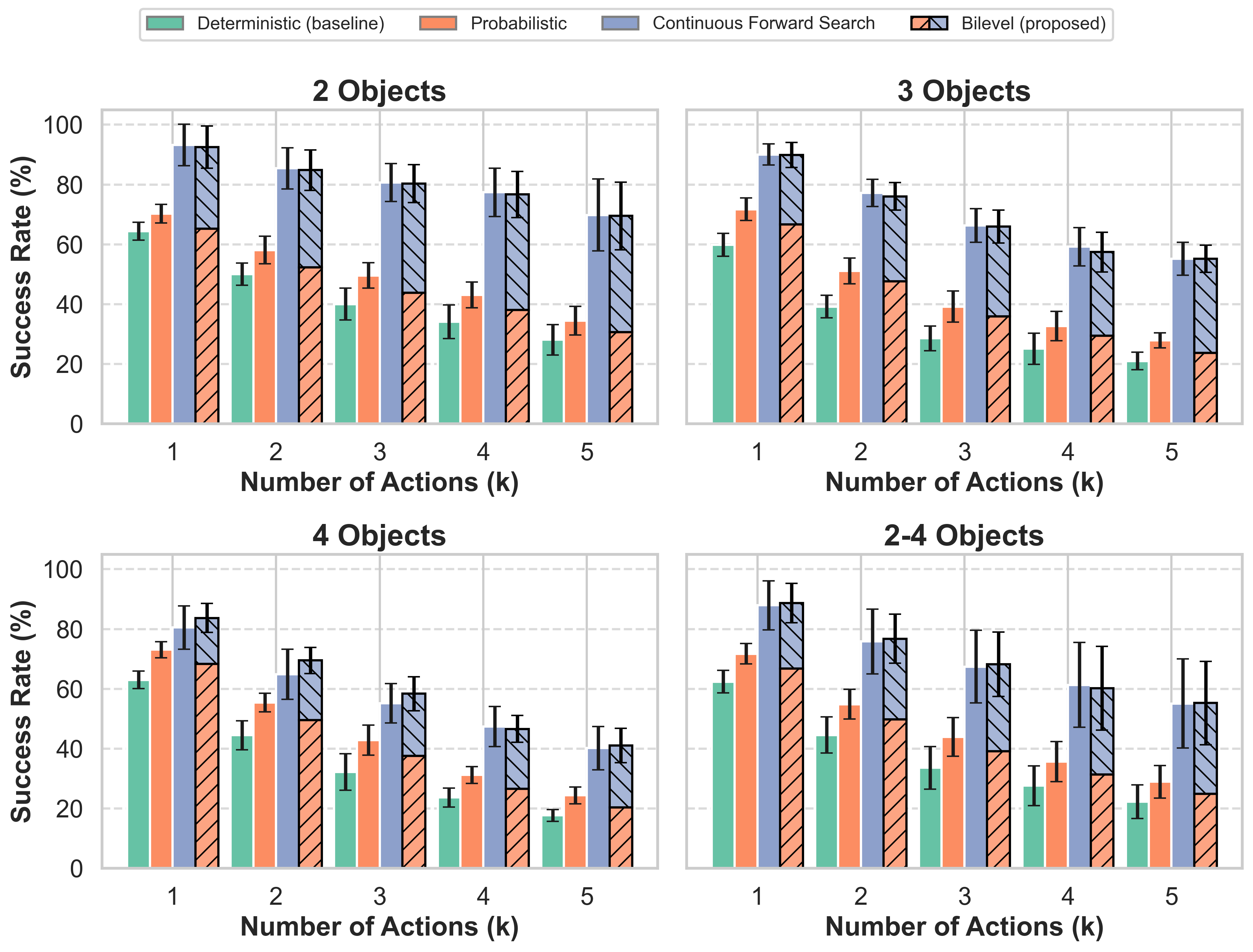}
   \caption{Planning success rates of all methods over ten runs. Hatched orange and blue bars show the Probabilistic Planner (Level 1) and Continuous Forward Search fallback (Level 2) contributions, respectively.}
\label{fig:planning_results}
\end{figure}

We compared the planning performance of four planning methods. Fig.~\ref{fig:planning_results} demonstrates the results, which report success rates across different numbers of objects ($n \in \{2,3,4\}$) and action sequence lengths ($k \in \{1,2,3,4,5\}$). As discussed in earlier sections, the Probabilistic Symbolic Planner enhanced performance relative to the Deterministic Symbolic Planner. Additionally, Continuous Forward Search performs better than both symbolic planning methods across all conditions, serving as the most powerful, albeit expensive, approach.

The difference in planning performance between the proposed Bilevel Planner and Continuous Forward Search was not statistically significant ($p = 0.21$), indicating that the bilevel planning approach performs comparably to the most capable method while also benefiting from symbolic reasoning and plan verification. Therefore, our approach preserves the efficiency of symbolic reasoning, ensures the robustness of the candidate plans through verification, and employs continuous forward search solely for challenging problems. In terms of efficiency, we measured the planning time of each method. As reported in Table~\ref{tab:planning_time}, the Bilevel Planner is overall faster than the Continuous Forward Search, and the Bilevel Planner’s relative speedup increases as the number of objects increases (from $\approx1.16$x at $n=2$, to $\approx1.21$x at $n=3$,~and $\approx1.81$x at $n=4$), indicating that the continuous search~space expands more rapidly as more objects are introduced into the problem. Also, the Bilevel Planner solved $\approx60\%$ of~the problems in Level 1, utilizing symbolic planning, with a mean planning time of $0.006 \pm 0.003$ s for these problems, whereas Continuous Forward Search averaged $0.932 \pm 0.566$ s for the same problems. However, for the remaining problems requiring the continuous fallback (Level 2), the Bilevel Planner averaged $5.95 \pm 1.42$~s versus $5.87 \pm 1.36$ s for Continuous Forward Search for the same problems, with the $\approx0.1$~s overhead stemming from the unverified symbolic search (during Level~1). The overall gain of the Bilevel Planner thus comes from using the $\approx155x$ faster symbolic planning (solving in Level~1) on a portion of problems that are relatively easy (indicated by lower planning times for Continuous Forward Search), and solving the more challenging problems with continuous fallback (solving in Level 2). The overhead of continuous predictions during verification is negligible, since the verifier processes only as many states as there are actions for each selected symbolic plan. Overall, the Bilevel Planner performs comparably to Continuous Forward Search while gaining substantial efficiency from symbolic planning.

\begin{table}[t]
\centering
\caption{Planning time (s) comparison of the Bilevel Planner (Ours) and Continuous Forward Search (CFS) across 50 problems, over ten runs and the number of actions ($k \in {\{1,2,3,4,5\}}$).
}
\label{tab:planning_time}
\begin{tabular}{@{}lccc@{}}
\toprule
\textbf{} & \textbf{n = 2} & \textbf{n = 3} & \textbf{n = 4}  \\
\midrule
Ours  & $\mathbf{0.83 \pm 0.15}$ & $\mathbf{2.63 \pm 0.79}$ & $\mathbf{4.75 \pm 3.10}$  \\
CFS & $0.96 \pm 0.14$ & $3.18 \pm 0.72$ & $8.58 \pm 4.88$ \\
\bottomrule
\end{tabular}
\end{table}

The verified and successful plans found by the Probabilistic Symbolic Planner (hatched orange bar) form a substantial part of the overall success of the Bilevel Planner. As shown in Fig.~\ref{fig:planning_results}, these verified plans achieve a higher success rate than the Deterministic Planner. While the probabilistic approach explores a broader range of outcomes, the verification process helps produce more reliable plans. Although the verification process may discard a valid plan, it still yields higher success with a Probabilistic Planner than with deterministic planning and additionally enhances the system's robustness.

We further evaluated a closed-loop variant of the Bilevel Planner that replans toward the goal by re-observing the resulting state after each action is executed, to assess whether replanning improves the success rate, especially at higher action counts, where success rates are lower. If the executed plan does not reach the goal, the agent executes its first action and replans from the newly observed state to the goal, repeating this for up to ($k$ - $1$) rounds. As shown in Table~\ref{tab:replanning}, replanning improves the success rate across all action counts, with an improvement over the baseline without replanning that grows with the horizon, from $0.09$ at $k=2$ to $0.12$ at $k=5$. Almost all recovered cases are solved by the continuous forward search, as giving the re-observed intermediate state instead of the predicted state mitigates the accumulation of prediction error along the horizon. Symbolic planning failures usually arise from the absence of a learned operator that covers the required transition and are therefore unlikely to be solved by replanning. Instead, they can be addressed through improved data collection and an exploration process that will mitigate coverage limitations.

\begin{table}[t]
\centering
\caption{Bilevel planning success rate with and without replanning (Rp.) across 50 problems ($k > 1$), over ten runs, and the number of objects ($n \in {\{2,3,4\}}$).
}
\label{tab:replanning}
\begin{tabular}{@{}lcccc@{}}
\toprule
\textbf{Bilevel} & \textbf{k = 2} & \textbf{k = 3} & \textbf{k = 4} & \textbf{k = 5} \\
\midrule
\textbf{w/o} Rp. & $0.82 \pm 0.03$ & $0.74 \pm 0.05$ & $0.67 \pm 0.07$ & $0.60 
\pm 0.04$ \\
\textbf{w} Rp.   & $\mathbf{0.91 \pm 0.02}$ & $\mathbf{0.85 \pm 0.02}$ & $\mathbf{0.80 \pm 0.04}$ & $\mathbf{0.72 \pm 0.05}$ \\
\bottomrule
\end{tabular}
\end{table}

%% file: limitations.tex
\section{Limitations}

The quality of the learned models directly determines the performance of both planning stages. Currently, the transition dataset collected via random exploration does not efficiently cover the rare but critical state transitions required for long-horizon planning, thereby impeding the quality of operator learning from the \textbf{Symbolic Model} and accurate effect prediction from the \textbf{Dynamics Model}. Active, model-guided exploration or curriculum learning, which would allow skills to be learned progressively as the number of actions and/or objects involved increases, could address this. The verifier also uses a fixed threshold, which could be made adaptive to problem features such as the number of objects. Additionally, the verifier could provide a high-quality starting point for the continuous forward search rather than starting from scratch.

%% file: conclusion.tex
\section{Conclusion}

We introduced a bilevel planning framework that successfully combines the efficiency of symbolic reasoning and the accuracy of continuous search. Our dual-model scheme uses a \textbf{Symbolic Model} to discover symbols and learn probabilistic operators for symbolic planning, and a distinct \textbf{Dynamics Model} to predict effects during plan verification and for planning in continuous state space. In experiments, we validated the dual-model design, showed that the probabilistic planner achieves significantly higher success rates than its deterministic counterpart, and the verification mechanism reliably validates symbolic plans. Finally, we demonstrated that the proposed bilevel framework performs comparably to a full Continuous Forward Search while retaining the computational benefits of symbolic reasoning, engaging the more expensive search only when required.